# Weighted Attribute Fusion Model for Face Recognition

S.Sakthivel

Assistant Professor, Department of Information
Technology
Sona college of Technology, Salem, India
sakthits@rediffmail.com

Dr.R.Lakshmipathi

Professor, Department of Electrical and Electronic
Engineering
St.Peter's Engineering College, Chennai, India
drrlakshmipathi@yahoo.com

*Abstract*—**Recognizing a face based on its attributes is an easy task for a human to perform as it is a cognitive process. In recent years, Face Recognition is achieved with different kinds of facial features which were used separately or in a combined manner. Currently, Feature fusion methods and parallel methods are the facial features used and performed by integrating multiple feature sets at different levels. However, this integration and the combinational methods do not guarantee better result. Hence to achieve better results, the feature fusion model with multiple weighted facial attribute set is selected. For this feature model, face images from predefined data set has been taken from Olivetti Research Laboratory (ORL) and applied on different methods like Principal Component Analysis (PCA) based Eigen feature extraction technique, Discrete Cosine Transformation (DCT) based feature extraction technique, Histogram Based Feature Extraction technique and Simple Intensity based features. The extracted feature set obtained from these methods were compared and tested for accuracy. In this work we have developed a model which will use the above set of feature extraction techniques with different levels of weights to attain better accuracy. The results show that the selection of optimum weight for a particular feature will lead to improvement in recognition rate.**

## I. INTRODUCTION

Face recognition is an important part of today's emerging biometrics and video surveillance markets. Face Recognition can benefit the areas of: Law Enforcement, Airport Security, Access Control, Driver's Licenses & Passports, Homeland Defense, Customs & Immigration and Scene Analysis. Face recognition has been a research area for almost 30 years, with significantly increased research activity since 1990[16] [15]. This has resulted in the development of successful algorithms and the introduction of commercial products. But, the researches and achievements on face recognition are still in its initial stages of development. Although face recognition is still in the research and development phase, several commercial systems are currently available and research organizations are working on the development of more accurate and reliable systems. Using the present technology, it is impossible to completely model human recognition system and reach its performance and accuracy. However, the human brain has its shortcomings in some aspects. The benefits of a computer system would be its capacity to handle large amount of data and ability to do a job in a predefined repeated manner. The observations and findings about human face recognition system will be a good starting point for automatic face attribute.

### A. Early Works

Face recognition has gained much attention in the last two decades due to increasing demand in security and law enforcement applications. Face recognition methods can be divided into two major categories, appearance-based method and feature-based method. Appearance-based method is more popular and achieved great success [3].

Appearance-based method uses the holistic features of a 2-D image [3]. Generally face images are captured in very high dimensionality, normally which is more than 1000 pixels. It is very difficult to perform face recognition based on original face image without reducing the dimensionality by extracting the important features. Kirby and Sirovich first used principal component analysis (PCA) to extract the features from face image and used them to represent human face image [16]. PCA seeks for a set of projection vectors which project the image data into a subspace based on the variation in energy. Turk and Pentland introduced the well-known Eigenface method [15]. Eigenface method incorporates PCA and showed promising results. Another well-known method is Fisher face. Fisher face incorporates linear discriminant analysis (LDA) to extract the most discriminant features and to reduce the dimensionality [3]. when it comes to solving problems of pattern classification, LDA based algorithms outperform PCA based ones, since the former optimizes the low dimensional representation of the objects with focus on the most discriminant feature extraction while the latter achieves simply object reconstruction [9][10][11]. Recently there has been a lot of interest in geometrically motivated approaches to data analysis in high dimensional spaces. This case is concerned with data drawn from sampling a probability distribution that has support on or near a sub manifold of Euclidean space [5].

Let us consider a collection of data points of n-dimensional real vectors drawn from an unknown probability distribution. In increasingly many cases of interest in machine learning and data mining, one is confronted with the situation which is very large. However, there might be reason to suspect that the "intrinsic dimensionality" of the data is much lower. This leads one to consider methods of dimensionality reduction [1][2][8] that allow one to represent the data in a lower dimensional space. A great number of dimensionality reduction techniques exist in the literature.

In practical situations, where data is prohibitively large, one is often forced to use linear (or even sub linear) techniques. Consequently, projective maps have been the subject of considerable investigation. Three classical yet popular forms of linear techniques are the methods of PCA [6] [14] [1][2], multidimensional scaling (MDS) [6] [14], and LDA [14] [11]. Each of these is an eigenvector method designed to model linear variability's in high-dimensional data. More recently, frequency domain analysis methods [3] such as discrete Fourier transform (DFT), discrete wavelet transform (DWT) and discrete cosine transform (DCT) have been widely adopted in face recognition. Frequency domain analysis methods transform the image signals from spatial domain to frequency domain and analyse the features in frequency domain. Only limited low-frequency components which contain high energy are selected to represent the image. Unlike PCA, frequency domain analysis methods are data independent [3]. They analyse image independently and do not require training images. Furthermore, fast algorithms are available for the ease of implementation and have high computation efficiency.

In [3] new parallel models for face recognition were presented. Feature fusion is one of the easy and effective ways to improve the performance. Feature fusion method is performed by integrating multiple feature sets at different levels. However, feature fusion method does not guarantee better result [3]. One major issue is feature selection. Feature selection plays a very important role to avoid overlapping features and information redundancy. New parallel model for face recognition utilizes information from frequency and spatial domains, addresses both features and processes in parallel way. It is well-known that image can be analysed in spatial and frequency domains. Both domains describe the image in very different ways. The frequency domain features [3] are extracted using techniques like DCT, DFT and DWT methods respectively. By utilizing these two or more very different features, a better performance is guaranteed.

Feature fusion method suffers from the problem of high dimensionality because of the combined features. It may also contain redundant and noisy data. PCA is applied on the features from frequency and spatial domains to reduce the dimensionality and extract the most discriminant information [3]. It is surprising that until recent study demonstrated that colour information makes contribution and enhances robustness in face recognition[4].

## II. MATERIALS AND METHODS

In statistics, dimension reduction is the process of reducing the number of random variables under consideration, and can be divided into feature selection and feature extraction.

### A. Extracting Eigen Features F1

In previous work [1] five algorithms are evaluated, namely PCA, Kernel PCA, LDA [9] [10], Locality Preserving Projections (LPP) [1] [8] and Neighbourhood Preserving Embedding (NPE) [1][2] [5] for dimensionality reduction and feature extraction and found that Kernel PCA was the best performer. This work uses PCA based algorithm to show some improvements in its performance due to the use of Multiple Weighted Facial Attribute Sets. The basic idea of PCA [5] is to project the data along the directions of maximal variances so that the reconstruction error can be minimized. Given a set of data points x1... xn, let a be the transformation vector and $yi = a^T x_i$. The objective function of PCA is as follows:

$$a_{opt} = \arg\max_a \sum_{i=1}^{n} (y_i - \overline{y})^2 = \arg\max_a a^T C_a \text{-------(1)}$$

In equation $\overline{y} = \frac{1}{n} \sum y_i$ and C is the data covariance the

eigen.The basic functions of PCA are the eigenvectors of the data covariance matrix corresponding to the largest eigenvalues.While PCA seeks direction that are efficient for representation.

### B. Extracting DCT Features F2

The DCT [2] can be used to Create DCT feature Set of the Face. The Discrete Cosine Transform is a real domain transform which represents the entire image as coefficients of different frequencies of cosines (which are the basis vectors for this transform). The DCT of the image is calculated by taking 8x8 blocks of the image in Figure 1, which is then transformed individually. The 2D DCT of an image gives the result matrix such that top left corner represents lowest frequency coefficient while the bottom right corner is the highest frequency.

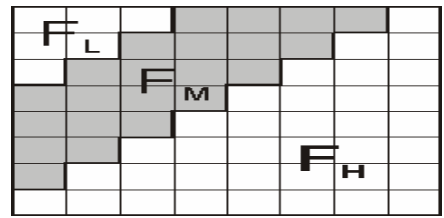

Figure 1 The frequency domain representation of an image

The 1-D *discrete cosine transform* (DCT) is defined as

$$C(u) = \alpha(u)\sum_{x=0}^{N-1} f(x) \cdot \cos\left[\frac{(2x+1)u\pi}{2N}\right] \quad \text{------- (2)}$$

Similarly, the inverse DCT is defined as

$$f(x) = \sum_{u=0}^{N-1} \alpha(u)C(u) \cdot \cos\left[\frac{(2x+1)u\pi}{2N}\right] \quad \text{------- (3)}$$

for x= 0,1,2,…,N  1 . In both **equations (2)** and **(3)**

*(u) is   defined as*

$$\alpha(u) = \begin{cases} \sqrt{1/N} & \text{for} \quad u = 0 \\ \sqrt{2/N} & \text{for} \quad u = 1,2,\ldots,N-1 \end{cases} \qquad \text{------ (4)}$$

The corresponding 2-D DCT, and the inverse DCT are defined as

$$C(u,v) = \alpha(u)\alpha(v)\sum_{x=0}^{N-1}\sum_{y=0}^{N-1} f(x,y)\cdot\cos\left[\frac{(2x+1)u\pi}{2N}\right]\cdot\cos\left[\frac{(2y+1)v\pi}{2N}\right] \text{----- (5)}$$

for $u,v = 0,1,2,\ldots,N-1$ and $\alpha(u)$ and $\alpha(v)$ are defined in **(4)**.

The inverse transform is defined as

$$f(x,y) = \sum_{u=0}^{N-1}\sum_{v=0}^{N-1}\alpha(u)\alpha(v)C(u,v)\cdot\cos\left[\frac{(2x+1)u\pi}{2N}\right]\cdot\cos\left[\frac{(2y+1)v\pi}{2N}\right]$$
$$\text{------(6)}$$

The advantage of DCT is that it can be expressed without complex numbers. The DCT transform Equation (5) can be expressed as separable, (like 2-D Fourier transform), i.e. it can be obtained by two subsequent 1-D DCT in the same way as Fourier transform. Equation (6) shows the Inverse transformation.

### C. The Histogram Feature Vector F3

The distribution of gray levels occurring in an image is called gray level histogram. It is a graph showing the frequency of occurrence of each gray level in the image versus the gray level itself. The plot of this function provides a global description of the appearance of the image. The histogram of a digital image with gray levels in the range [0, L-1] is a discrete function .

$$P(r_k) = n_k / n \qquad \text{------- (7)}$$

Where,

$r_k$ → Kth gray level

$n_k$ → No of pixels in the image with that gray level.

$n$ → total number of pixels in the image.

$K = 0, 1, 2\ldots L\text{-}1.$

$L = 256.$ (For 256 level gray images).

In Equation 7 $P(r_k)$ gives an estimate of the probability of occurrence of gray level $r_k$. If we use L value of small size, then $n_k$ will contain a range of nearest values in L number f bins. So for constructing Histogram Based Feature, the set $n_k$ and the mid values of the bin $n_k$ were combined.

### D. The Intensity Based Feature F4

Intensity Feature set can be formed by using values like mean Median Mode of the 256 gray level images. We can use one or many of this values to represent the average intensity of the face image.

### E. Neural Networks and Learning Paradigms

In principle, the popular neural network can be trained to recognize face images directly. However, a simple network can be very complex and difficult to train [12][17]. There are three major learning paradigms, each corresponding to a particular abstract learning task. These are supervised learning, unsupervised learning and reinforcement learning. Usually any given type of network architecture can be employed in any of those tasks.

### F. Learning Algorithms

Training a neural network model essentially means selecting one model from the set of allowed models (or, in a Bayesian framework, determining a distribution over the set of allowed models) that minimizes the cost criterion. There are numerous algorithms available for training neural network models; most of them can be viewed as a straightforward application of optimization theory and statistical estimation. Most of the algorithms used in training artificial neural networks are employing some form of gradient descent. This is done by simply taking the derivative of the cost function with respect to the network parameters and then changing those parameters in a gradient-related direction. Evolutionary methods simulated annealing, and Expectation-maximization and non-parametric methods are among other commonly used methods for training neural networks.

### G. Support vector machines (SVMs)

Support vector machines are a set of related supervised learning methods used for classification and regression. Viewing input data as two sets of vectors in an n-dimensional space, an SVM will construct a separating hyper plane in that space, one which maximizes the margin between the two data sets[7][13]. To calculate the margin, two parallel hyper planes are constructed, one on each side of the separating hyper plane, which is "pushed up against" the two data sets. Intuitively, a good separation is achieved by the hyper plane that has the largest distance to the neighbouring data points of both classes, since in general the larger the margin the better the generalization error of the classifier. For the linearly separable case, a hyper-plane separating the binary decision classes in the three-attribute case can be represented as the following equation:

$$y = w_0 + w_1 x_1 + w_2 x_2 + w_3 x_3 \qquad \text{------ (8)}$$

In Equation (8), $y$ is the outcome $x_i$, are the attribute values, and there are four weights $w_i$ to be learned by the learning algorithm. In the above equation, the weights $w_i$ are parameters that determine the hyper-plane. The maximum margin hyper-plane can be represented as the following equation in terms of the support vectors:

$$y = b + \sum \alpha_i y_i K(x(t).x) \qquad \text{------ (9)}$$

In Equation (9) the function $K(x(t).x)$ is defined as the kernel function. There are different kernels for generating the inner products to construct machines with different types of nonlinear decision surfaces in the input space.

SVM is selected as the classifying function. One distinctive advantage this type of classifier has over traditional neural networks is that SVMs can achieve better generalization performance. Support vector machine is a pattern classification algorithm developed by Vapnik [13]. It is a binary classification method that finds the optimal linear decision surface based on the concept of structural risk minimization. As shown by Vapnik, this maximal margin decision boundary can achieve optimal worst-case generalization performance. Note that SVMs are originally designed to solve problems where data can be separated by a linear decision boundary

### III. THE MODEL OF PROPOSED SYSTEM

Given a set of feature vectors belonging to n classes, a Support Vector Machine (SVM) finds the hyper plane that separates the largest possible fraction of features of the same classes on the corresponding space, while maximizing the distance from either class to the hyper plane. Generally a suitable transformation is first used to extract features of face images and then discrimination functions between each class of images are learned by SVMs. Figure 2 shows the feature set creation for face images from ORL database whereas Figure 3 shows the architecture for training and testing the SVM with weighted attribute set.

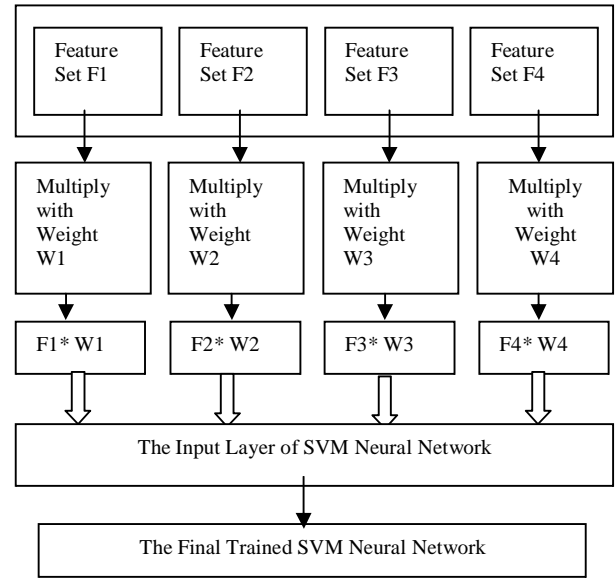

Figure 3: Architecture for Training and Testing the SVM with Weighted Attribute Set

#### A. Steps Involved in Training

1) Load a set of 'n' ORL Face Images For Training

2) Resize the Images in to 48x48 pixel size to reduce the memory requirements of the overall application

3) Reshape the Images in to 1x 2304 and prepare an n x 2304 Feature Matrix representing the training data set.

4) Apply a Feature Extraction, DCT, Histogram and Dimensionality Reduction technique and find the Four of above said Features sets F1, F2, F3 and F4.

5) Multiply the Weights W1, W2, W3 and W4 with F1, F2, F3 and F4.

6) Create an SVM network with "f1+f2+f3+f4" inputs where f1, f2, f3 and f4 are the corresponding feature lengths.

7) Train a SVM using the Weighted Feature Set.

#### B. Steps Involved in Testing

1) The first three steps of the above procedures will be repeated with test image set of ORL database to obtain a Feature Matrix representing the testing data set.

2) Project the matrix using Previous Eigen Vector Matrix and find the input Eigen Feature iF1.

3) Similarly find other input feature sets iF2, iF3 and iF4 of the input Image.

4) Classify the feature set [iF1 iF2 iF3 iF4] with previously trained SVM network.

5) Calculate Accuracy of Classification

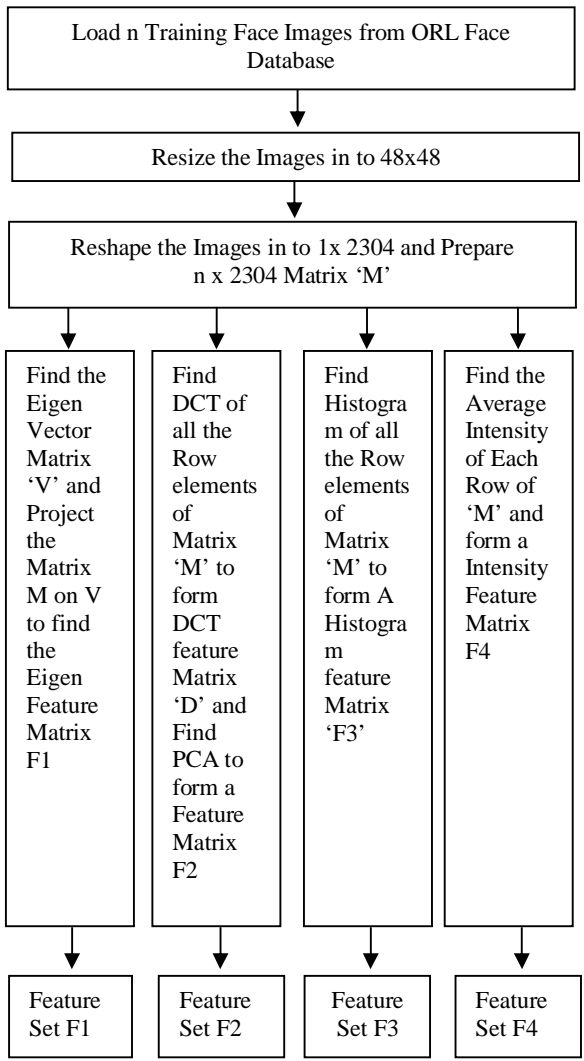

Figure 2: The Feature Set Creation

## IV.  IMPLEMENTATION RESULTS AND ANALYSIS

The performance of proposed face recognition model was tested with the standard set of images called "ORL Face Database". The ORL Database of Faces contains a set of face images used in the context of a face recognition project carried out in collaboration with the Speech, Vision and Robotics Group of the Cambridge University Engineering Department. There are ten different images, each of 40 distinct subjects. For some subjects, the images were taken at different times, varying the lighting, facial expressions (open / closed eyes, smiling / not smiling) and facial details (glasses / no glasses). All the images were taken against a dark homogeneous background with the subjects in an upright, frontal position (with tolerance for some side movement).

Set of images from ORL databases were used for Training and Testing. The accuracy of recognition with multiple weighted attribute sets as well as a single attribute sets has been evaluated. The following table shows the overall results of these two types of techniques with different number of input face images.

TABLE 1  ACCURACY OF RECOGNITION WITH SINGLE ATTRIBUTE SET

| No. of Faces used for Training and Testing | Accuracy of Recognition with different Weight Sets (%) | | | |
|---|---|---|---|---|
| | W1=1 W2=0 W3=0 W4=0 | W1=0 W2=1 W3=0 W4=0 | W1=0 W2=0 W3=1 W4=0 | W1=0 W2=0 W3=0 W4=1 |
| 10 | 90.00 | 40.00 | 80.00 | 90.00 |
| 20 | 80.00 | 25.00 | 60.00 | 65.00 |
| 30 | 83.33 | 16.67 | 56.67 | 76.67 |
| 40 | 80.00 | 12.50 | 52.50 | 70.00 |
| Average | **83.33** | **23.54** | **62.29** | **75.42** |

TABLE 2 ACCURACY OF RECOGNITION WITH MULTIPLE ATTRIBUTE SET

| No. of Faces used for Training and Testing | Accuracy of Recognition with different Weight Sets (%) | | | | |
|---|---|---|---|---|---|
| | W1=1 W2=1 W3=0 W4=0 | W1=.5 W2=1 W3=0 W4=0 | W1=.5 W2=1 W3=0 W4=1 | W1=1 W2=1 W3=1 W4=1 | W1=.12 W2=0 W3=1 W4=0 |
| 10 | 90.00 | 100 | 100 | 100 | 100 |
| 20 | 80.00 | 85.00 | 85.00 | 85.00 | 90.00 |
| 30 | 86.67 | 86.67 | 86.67 | 83.33 | 86.67 |
| 40 | 82.50 | 85.00 | 82.50 | 82.50 | 82.50 |
| Average | **84.79** | **89.17** | **88.54** | **87.71** | **89.79** |

As shown in the above table and the following graphs, the performance of recognition while using Single Attribute Set as well as multiple attribute sets with same priority or weight will lead to poor recognition results.  For example, all the results of **Table 1** which used a single attribute at a time for recognition, is in some what poor than **Table 2** which is using combined multi attributes. Further, if we note the fourth row (40 images) corresponding to the weights (W1=1, W2=1, W3=1, W4=1) which is using all the attributes with same weight, the result is poor while comparing it with column 2 (W1=0.5, W2=1, W3=0, W4=0). The following Graphs show the performance of the two different approaches. In **Figure 4** Line charts shows the Performance with single attribute set, but in **Figure 5** Line charts shows the performance with multiple weighted attribute sets.

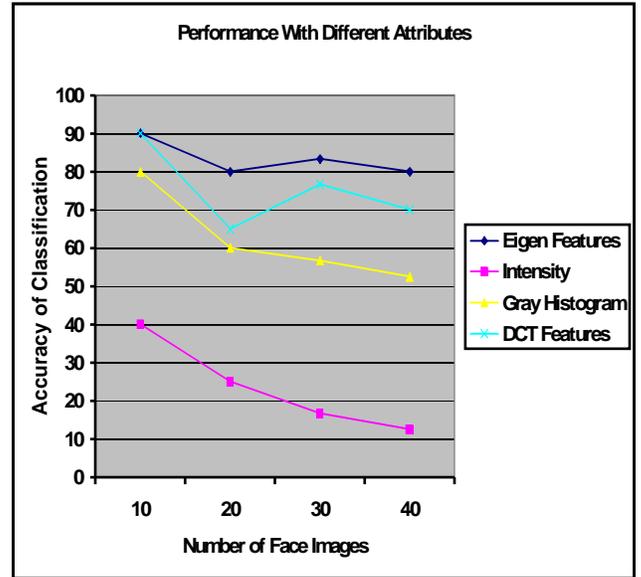

Figure 4 Performance with single attribute

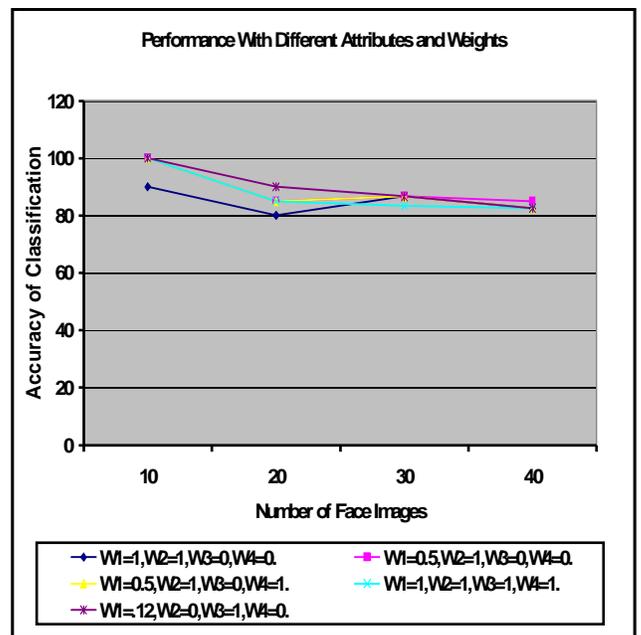

Figure 5 Performance with multiple weighted attribute sets

The Following Two Charts shows the average performance of the two approaches. In Figure 6, column charts shows the average performance with single attribute set.

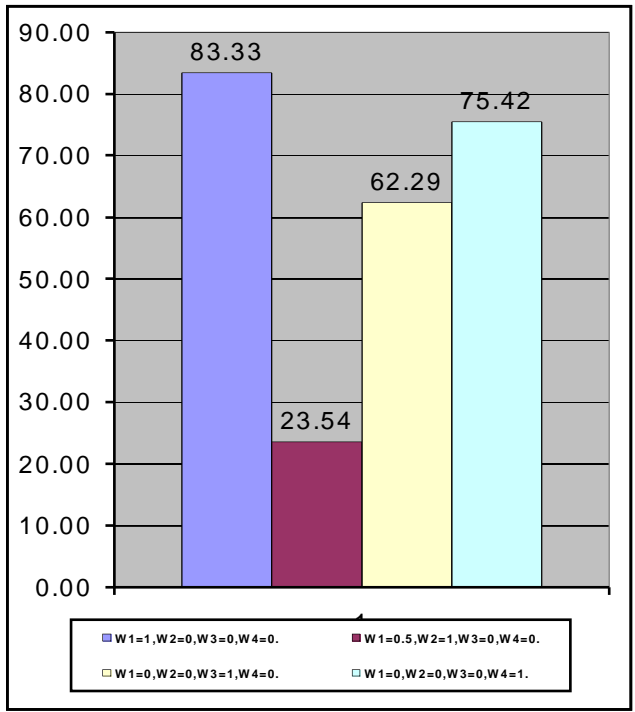

Figure 6 average performance with single attribute set

In **Figure 7,** column charts shows the average performance with multiple attribute sets.

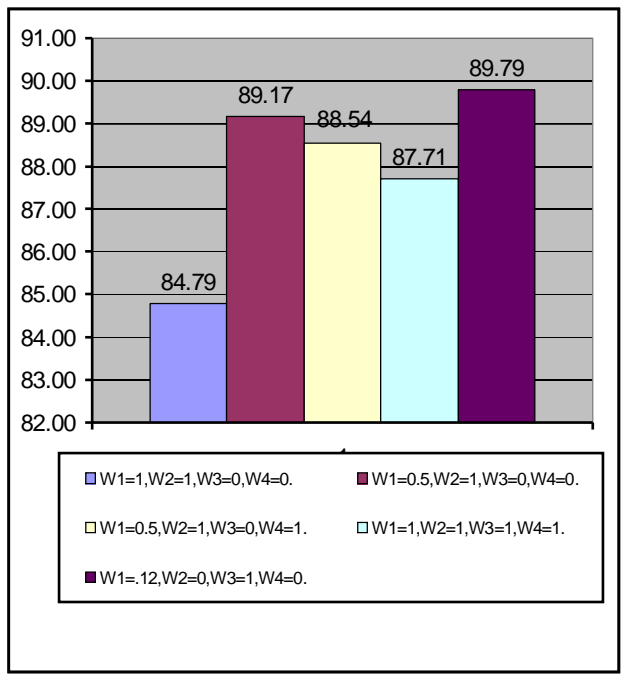

Figure 7 Average performances with multiple attribute set

## V. CONCLUSION

Complicated Face recognition techniques constantly facing very challenging and difficult problems in several research works. In spite of the great work done in the last 30 years, it is sure that the face recognition research community will have to work for at least the next few decades to completely solve the problem. Strong and coordinated efforts between the computer visions signal processing and psychophysics and neurosciences community is needed. The proposed Multiple Weighted Feature Attribute Sets based training provided significant improvement in terms of performance accuracy of the face recognition system. With ORL data set, a significant 5% performance improvement was observed during various tests. In this work, we have selected PCA as the main feature extraction technique. The weights of the used feature sets were decided based on trial and error method. This will be applicable for systems with predefined data sets. In future works, one may explore different techniques like Kernel PCA, LDA for better performance. So, future works may address methods for automatic estimation of the weights of the feature sets with respect to the application.

## REFERENCES

[1] S.Sakthivel, Dr.R.Lakshmipathi and M.A.Manikandan "Evaluation of Feature Extraction and Dimensionality Reduction Algorithms for Face Recognition using ORL Database", The Paper published in Proceedings of The 2009 International Conference on Image Processing, Computer Vision, and Pattern Recognition (IPCV'09) , Las Vegas, USA, 2009. ISBN: 1-60132-117-1, 1-60132-118-X (1-60132-119-8) Copyright 2009 CSREA Press.

[2] S.Sakthivel, Dr.R.Lakshmipathi and M.A.Manikandan "Improving the performance of machine learning based face recognition algorithm with Multiple Weighted Facial Attribute Sets" The paper published in proceedings of The Second International Conference on the Applications of Digital Information and Web Technologies, 2009 Volume, Issue, 4-6 Aug. 2009 Page(s):658 – 663, ISBN: 978-1-4244-4456-4, INSPEC Accession Number: 10905880, Digital Object Identifier: 10.1109/ICADIWT.2009.5273884

[3] Heng Fui Liau, Kah Phooi Seng, Li-Minn Ang and Siew Wen Chin, "New Parallel Models for Face Recognition" University of Nottingham Malaysia Campus, Malaysia - Recent Advances in Face Recognition, Published by In-Teh, 2008 Recent Advances in Face Recognition, Book edited by: Kresimir Delac, Mislav Grgic and Marian Stewart Bartlett, ISBN 978-953-7619-34-3, pp. 236, December 2008, I-Tech, Vienna, Austria.

[4] Khalid Youssef and Peng-Yung Woo, "A Novel Approach to Using Color Information in Improving Face Recognition Systems Based on Multi-Layer Neural Networks", Northern Illinois University USA - Recent Advances in Face Recognition, Published by In-Teh, 2008, Source: Recent Advances in Face Recognition, Book edited by: Kresimir Delac, Mislav Grgic and Marian Stewart Bartlett, ISBN 978-953-7619-34-3, pp. 236, December 2008, I-Tech, Vienna, Austria

[5] Xiaofei He; Deng Cai; Shuicheng Yan; Hong-Jiang Zhang,"Neighborhood preserving embedding", Tenth IEEE International Conference on Computer Vision, Volume 2, Issue, 17-21 Oct. 2005. Proceedings of the Tenth IEEE International Conference on Computer Vision (ICCV'05) 1550-5499/05

[6] Shuicheng Yan; Dong Xu; Benyu Zhang; Hong-Jiang Zhang,"Graph embedding: a general framework for dimensionality reduction", IEEE Computer Society Conference on Computer Vision and Pattern Recognition, Volume 2, June 2005. Proceedings of the 2005 IEEE Computer Society Conference on Computer Vision and Pattern Recognition (CVPR'05) 1063-6919/05.

[7] S. Canu, Y. Grandvalet, V. Guigue, and A. Rakotomamonjy. (2005). SVM and kernel methods Matlab toolbox Perception Systems at Information http://asi.insa-uen.fr/~arakotom/toolbox/index.html

[8] Xiaofei He, Partha Niyogi, "Locality Preserving Projections (LPP)", Computer Science Department, The University of Chicago, In Advances in Neural Information Processing Systems, Volume 16, page 37, Cambridge, MA, USA 2004, The MIT press. http://books.nips.cc/papers/files/nips16/NIPS2003_AA20.pdf

[9] Lu, J.; Plataniotis, K.N. & Venetsanopoulos, A. N. (2003) Face Recognition Using LDA based Algorithm", *IEEE trans.Neural Network*, vol.14, No 1, pp.195-199, January 2003. Digital Object Identifier 10.1109/ TNN.2002.806647.

[10] Yu, Hu. & Yang, J. (2001) A Direct LDA algorithm for high-dimension data with application to face recognition, *Pattern Recognition*, vol.34, pp. 2067-2070, 2001. DOI: 10.1016/S0031-3203(00)00162-X

[11] Chen, L.F.; Mark Liao, H.Y.; Ko, M.T.; Lin, J.C. & Yu, G.J. (2000) A new LDA-based face recognition system which can solve the small space size problem, *Pattern Recognition*, vol.33, pp.1703-1726, 2000. DOI: 10.1016/ S0031-3203(99)00139-9

[12] H.A. Rowley and T. Kanade, "Neural network-based face detection", IEEE Trans. on PAMI, vol. 20, no. 1, pp. 23-38,Jan1998.http://www.informedia.cs.cmu.edu/documents/ rowley-ieee.pdf

[13] E. Osuna, R. Freund, and F. Girosi, "Training support vector machines: an application to face detection," in Proc. IEEE Conf. Computer Vision Pattern Recognition, 1997. http://doi.ieeecomputersociety.org/10.1109/CVPR.1997.609310

[14] Imola K. Fodor, Center for Applied Scientific Computing, Lawrence Livermore National Laboratory "A survey of dimension reduction techniques"June2002. https://computation.llnl.gov/casc/sapphire/pubs/148494.pdf

[15] Turk, M. & Pentland, A. (1991) Eigenfaces for recognition, *Journal of Cognitive Neuroscience*, vol. 3, no. 1, pp. 71–86, Mar 1991. http://portal.acm.org/ citation.cfm?id=1326894

[16] Kirby, M. & Sirovich, L. (1990) Application of the Karhunen-Loeve procedure of the characteristic of human faces, *IEEE Trans. Pattern Anal. Machine Intell*, vol.12, pp 103-108, Jan, 1990. DOI 10.1109/34.41390

[17] Shang-Hung Lin, "An Introduction to Face Recognition Technology", Information science special issue on multimedia informing Technologies - part2 volume 3 No 1, 2000. http://inform.nu/Articles/Vol3/v3n1p01-07.pdf

## AUTHORS PROFILE

[1] **S. Sakthivel** received his M.E Computer science from Sona College of Technology, Affiliated to Anna University, Chennai, India in the year 2004.Currently, pursuing his Ph.D., in Anna University, Chennai,Tamilnadu.He has a work experience of 10 years. At present working as a Assistant Professorin the department of Information Technology. He has published paper in international journal.He has participated and presented research papers in various national and international seminars and conferences. He is an Life member of ISTE.

[2] **Dr.R.Lakshimpathi** received the B.E degree in 1971 and M.E degree in 1973 from College of Engineering, Guindy, and Chennai.He received his PhD degree in High Voltage Engineering from Indian Institute of Technology, Chennai, India. He has 36 years of teaching experience in various Government Engineering Colleges in Tamilnadu and he retired as Principal and Regional Research Director at Alagappa Chettiar College of Engineering and Technology, Karaikudi.He is now working as Professor in Electrical and Electronics Engineering department at St.Peters Engineering College, Chennai.His areas of research include HVDC Transmission, Power System Stability and Electrical Power Semiconductor Drives.